\documentstyle[aclap]{article}

\newcommand     {\Datr}             {{\sf DATR}}
\newcommand     {\node}[1]          {\mbox{\bf #1}}
\newcommand     {\path}[1]          {\langle #1 \rangle}
\newcommand     {\val}[1]           {\mbox{\bf #1}}
\newcommand     {\npp}[2]           {\node{#1}:\path{#2}}
\newcommand     {\defeq}[2]         {#1 == #2}
\newcommand     {\exteq}[2]         {#1 = #2}
\newcommand     {\quoted}[1]        {\mbox{``}#1\mbox{''}}
\newcommand     {\Node}             {\mbox{\sc node}}
\newcommand     {\Atom}             {\mbox{\sc atom}}
\newcommand     {\Desc}             {\mbox{\sc desc}}
\newcommand	{\Theory}	    {\mbox{$\cal T$}}
\newcommand     {\denotes}[1]       {[\![ #1 ]\!]}

\newtheorem{Def}{Definition}[section]

\makeatletter

\@ifundefined{LaTeXe}{}{
   \let\normalsize\@normalsize
}

\makeatother

\author{Bill Keller \\ School
of Cognitive and Computing Sciences \\ The University of Sussex \\ Brighton,
UK \\
email: billk@cogs.susx.ac.uk}
\title{\Datr\ Theories and \Datr\ Models}

\begin{document}

\maketitle
\bibliographystyle{acl}

\begin{abstract}

Evans and Gazdar \cite{eg89a,eg89b} introduced \Datr\ as a simple,
non-monotonic language for representing natural language lexicons.
Although a number of implementations of \Datr\ exist, the full language
has until now lacked an explicit, declarative semantics. This paper
rectifies the situation by providing a mathematical semantics for
\Datr. We present a view of \Datr\ as a language for defining certain
kinds of partial functions by cases. The formal model provides a
transparent treatment of \Datr 's notion of global context. It is shown
that \Datr 's default mechanism can be accounted for by interpreting
value descriptors as families of values indexed by paths.

\end{abstract}

\section{Introduction}
\label{sec-intro}

\Datr\ was introduced by Evans and Gazdar \shortcite{eg89a,eg89b} as a
simple, declarative language for representing lexical knowledge in terms
of path/value equations.  The language lacks many of the constructs
found in general purpose, knowledge representation formalisms, yet it has
sufficient expressive power to capture concisely the structure of
lexical information at a variety of levels of linguistic description.
At the present time, \Datr\ is probably the most widely-used formalism
for representing natural language lexicons in the natural language
processing (NLP) community. There are around a dozen different
implementations of the language and large \Datr\ lexicons have been
constructed for use in a variety of applications
\cite{ce90,andr92,cahill94}. \Datr\ has been applied to problems in
inflectional and derivational morphology \cite{gazd92,kilb92,cf93},
lexical semantics \cite{kilg93}, morphonology \cite{cahill93}, prosody
\cite{gb91} and speech \cite{andr92}. In more recent work, the language
has been used to provide a concise encoding of Lexicalised Tree
Adjoining Grammar \cite{egw94,egw95}.

A primary objective in the development of \Datr\ has been the provision
of an explicit, mathematically rigorous semantics. This goal was
addressed in one of the first publications on the language
\cite{eg89b}. The definitions given there deal with a subset of \Datr\
that includes core features of the language such as the notions of local
and global inheritance and \Datr's default mechanism. However, they
exclude some important and widely-used constructs, most notably string
(or `list') values and evaluable paths. Moreover, it is by no means
clear that the approach can be generalized appropriately to cover these
features.  In particular, the formal apparatus introduced by Evans and
Gazdar in \shortcite{eg89b} provides no explicit model of \Datr\ 's
notion of {\em global context\/}. Rather, local and global inheritance
are represented by distinct semantic functions ${\cal L}$ and
${\cal G}$. This approach is possible only on the (overly restrictive)
assumption that \Datr\ statements involve either local or global
inheritance relations, but never both.

The purpose of the present paper is to remedy the deficiencies of the
work described in \cite{eg89b} by furnishing \Datr\ with a transparent,
mathematical semantics. There is a standard view of \Datr\ as a language
for representing a certain class of non-monotonic inheritance networks
(`semantic nets').  While this perspective provides an intuitive and
appealing way of thinking about the structure and representation of
lexical knowledge, it is less clear that it provides an accurate or
particularly helpful picture of the \Datr\ language itself. In fact,
there are a number of constructs available in \Datr\ that are impossible
to visualize in terms of simple inheritance hierarchies. For this
reason, the work described in this paper reflects a rather different
perspective on \Datr, as a language for defining certain kinds of
partial functions by cases. In the following sections this viewpoint is
made more precise. Section~\ref{sec-theo} presents the syntax of the
\Datr\ language and introduces the notion of a \Datr\ theory.  An
informal introduction to the \Datr\ language is provided, by example, in
section~\ref{sec-overview}.  The semantics of \Datr\ is then covered in
two stages. Section~\ref{sec-interp} introduces \Datr\ interepretations
and describes the semantics of a restricted version of the language
without defaults. The treatment of implicit information is covered in
section~\ref{sec-default}, which provides a definition of a default
model for a \Datr\ theory.

\section{\Datr\ Theories}
\label{sec-theo}

Let $\Node$ and $\Atom$ be disjoint sets of symbols (the {\em nodes\/}
and {\em atoms\/} respectively). Nodes are denoted by $N$ and atoms by
$a$. The set \Desc\ of \Datr\ {\em value descriptors\/} (or simply {\em
descriptors\/}) is built up from the atoms and nodes as shown
below. Descriptors are denoted by $d$.

\begin{itemize}
\item $a \in \Desc$ for any $a \in \Atom$
\item For any $N \in \Node$ and $d_1 \ldots d_n \in \Desc$:
\begin{itemize}
\item[] $N:\langle d_1 \cdots d_n \rangle \in \Desc$
\item[] $\quoted{N:\langle d_1 \cdots d_n \rangle} \in \Desc$
\item[] $\quoted{\langle d_1 \cdots d_n \rangle} \in \Desc$
\item[] $\quoted{N}\in \Desc$
\end{itemize}
\end{itemize}

Value descriptors are either atoms or {\em inheritance descriptors\/},
where an inheritance descriptor is further distinguished as either {\em
local\/} (unquoted) or {\em global\/} (quoted). There is just one kind
of local descriptor (node/path), but three kinds of global descriptor
(node/path, path and node) \footnote{The syntax presented in
\cite{eg89a,eg89b} permits nodes and paths to stand as local
descriptors.  However, these additional forms can be viewed as
conventional abbreviations, in the appropriate syntactic context, for
node/path pairs}.

A {\em path\/} $\langle a_1 \ldots a_n \rangle$ is a (possibly empty)
sequence of atoms enclosed in angle brackets. Paths are denoted by $P$.
For $N$ a node, $P$ a path and $\alpha \in \Atom^*$ a (possibly empty)
sequence of atoms, an equation of the form $\exteq{N:P}{\alpha}$ is
called an {\em extensional sentence\/}.  Intuitively, an extensional
sentence $\exteq{N:P}{\alpha}$ states that the value associated with the
path $P$ at node $N$ is $\alpha$. For $\phi$ a (possibly empty) sequence
of value descriptors, an equation of the form $\defeq{N:P}{\phi}$ is
called a {\em definitional sentence\/}. A definitional sentence
$\defeq{N:P}{\phi}$ specifies a property of the node $N$, namely that
the path $P$ is associated with the value defined by the sequence of
value descriptors $\phi$.

A collection of equations can be used to specify the properties of
different nodes in terms of one another, and a finite set of \Datr\
sentences $\Theory$ is called a \Datr\ {\em theory\/}.  In principle, a
\Datr\ theory $\Theory$ may consist of any combination of \Datr\
sentences, either definitional or extensional, but in practice, \Datr\
theories are more restricted than this.  The theory $\Theory$ is said to
be {\em definitional\/} if it consists solely of definitional sentences
and it is said to be {\em functional\/} if it meets the following
condition:

\[
\defeq{N:P}{\phi} \mbox{ and }
\defeq{N:P}{\psi} \in \Theory
\mbox{ implies } \phi = \psi
\]

There is a pragmatic distinction between definitional and extensional
sentences akin to that drawn between the language used to define a database
and that used to query it.  \Datr\ interpreters conventionally treat
all extensional sentences as `goal' statements, and evaluate them as soon
as they are encountered. Thus, it is not possible, in practice, to
combine definitional and extensional sentences within a
theory\footnote{It is not clear why one would wish to do this anyway, but the
possibility is explicitly left open in the original definitions of
\cite{eg89a}.}. Functionality for \Datr\ theories, as defined above, is
really a syntactic notion. However, it approximates a deeper, semantic
requirement that the nodes should correspond to (partial) functions from
paths to values.

In the remainder of this paper we will use the term (\Datr) {\em
theory\/} always in the sense {\em functional, definitional\/} (\Datr)
{\em theory\/}.  For a given \Datr\ theory $\Theory$ and node $N$ of
$\Theory$, we write $\Theory/N$ to denote that subset of the sentences in
$\Theory$ that relate to the node $N$. That is:
\begin{eqnarray*}
\Theory/N & = & \{s \in \Theory | s = \defeq{N:P}{\phi} \}
\end{eqnarray*}
The set $\Theory/N$ is referred to as the {\em definition of\/} $N$
({\em in\/} $\Theory$).

\section{An Overview of \Datr}
\label{sec-overview}

An example of (a fragment of) a \Datr\ theory is shown in
figure~\ref{datreg-fig}.  The theory makes use of some standard
abbreviatory devices that enable nodes and/or paths to be omitted in
certain cases. For example, sets of sentences relating to the same node
are written with the node name implicit in all but the first-given
sentence in the set. Also, we write $\defeq{\npp{See}{}}{\node{Verb}}$
to abbreviate the definitional sentence
$\defeq{\npp{See}{}}{\npp{Verb}{}}$, and similarly elsewhere.

\begin{figure*}[t]
\[
\begin{array}{ll}
\node{Verb}: & \path{\val{syn cat}} == \val{verb} \\
 & \path{\val{syn type}} == \val{main} \\
 & \path{\val{mor form}} ==
   \quoted{\path{\val{mor }\quoted{\path{\val{syn form}}}}} \\
 & \path{\val{mor pres}} == \quoted{\path{\val{mor root}}} \\
 & \path{\val{mor past}} == \quoted{\path{\val{mor root}}}\mbox{ } \val{ed} \\
 & \path{\val{mor pres part}} ==
\quoted{\path{\val{mor root}}}\mbox{ }\val{ing} \\
 & \path{\val{mor pres sing three}} ==
   \quoted{\path{\val{mor root}}} \mbox{ }\val{s} \\
 & \\
\node{EnVerb}: & \path{} == \node{Verb} \\
 & \path{\val{mor past part}} ==
   \quoted{\path{\val{mor root}}} \mbox{ } \val{en} \\
 & \\
\node{Aux}: & \path{} == \node{Verb} \\
 & \path{\val{syn type}} == \val{aux} \\
 & \\
\node{Modal}:  & \path{} == \node{Aux} \\
 & \path{\val{mor pres sing three}} == \quoted{\path{\val{mor root}}} \\
 & \\
\node{Walk}: & \path{} == \node{Verb} \\
 & \path{\val{mor root}} == \val{walk} \\
 & \\
\node{Mow}: & \path{} == \node{EnVerb} \\
 & \path{\val{mor root}} == \val{mow} \\
 & \\
\node{Can}: & \path{} == \node{Modal} \\
 & \path{\val{mor root}} == \val{can}  \\
 & \path{\val{mor past}} == \val{could}
\end{array}
\]

\caption{A \Datr\ Theory}\label{datreg-fig}
\end{figure*}

The theory defines the properties of seven nodes: an abstract
$\node{Verb}$ node, nodes $\node{EnVerb}$, $\node{Aux}$ and
$\node{Modal}$, and three abstract lexemes $\node{Walk}$, $\node{Mow}$
and $\node{Can}$.  Each node is associated with a collection of
definitional sentences that specify values associated with different
paths. This specification is achieved either {\em explicitly\/}, or {\em
implicitly\/}. Values given explicitly are specified either {\em
directly\/}, by exhibiting a particular value, or {\em indirectly\/}, in
terms of local and/or global inheritance. Implicit specification is
achieved via \Datr\ 's default mechanism.

For example, the definition of the $\node{Verb}$ node gives the values
of the paths $\path{\val{syn cat}}$ and $\path{\val{syn type}}$
directly, as $\val{verb}$ and $\val{main}$, respectively.  Similarly,
the definition of $\node{Walk}$ gives the value of $\path{\val{mor
root}}$ directly as $\val{walk}$. On the other hand, the value of the
empty path at $\node{Walk}$ is given indirectly, by local inheritance,
as the value of the empty path at $\node{Verb}$.  Note that in itself,
this might not appear to be particularly useful, since the theory does
not provide an explicit value for the empty path in the definition of
$\node{Verb}$.  However, \Datr's default mechanism permits any
definitional sentence to be applicable not only to the path specified in
its left-hand-side, but also for any rightward extension of that path
for which no more specific definitional sentences exist.  This means
that the statement $\defeq{\npp{Walk}{}}{\npp{Verb}{}}$ actually
corresponds to a class of {\em implicit\/} definitional sentences, each
obtained by extending paths on the left- and the right-hand-sides of the
equation in the same manner. Examples include the following:

\[
\begin{array}{l}
\defeq{\npp{Walk}{\val{mor}}}{\npp{Verb}{\val{mor}}} \\
\defeq{\npp{Walk}{\val{mor form}}}{\npp{Verb}{\val{mor form}}} \\
\defeq{\npp{Walk}{\val{syn cat}}}{\npp{Verb}{\val{syn cat}}}
\end{array}
\]

Thus, the value associated with $\path{\val{syn cat}}$ at $\node{Walk}$
is given (implicitly) as the value of $\path{\val{syn cat}}$ at
$\node{Verb}$, which is given (explicitly) as $\val{verb}$. Also, the
values of $\path{\val{mor}}$ and $\path{\val{mor form}}$, amongst many
others, are inherited from $\node{Verb}$. In the same way, the value of
$\path{\val{syn cat}}$ at $\node{Mow}$ is inherited locally from
$\node{EnVerb}$ (which in turn inherits locally from $\node{Verb}$) and
the value of $\path{\val{syn cat}}$ at $\node{Can}$ is inherited locally
from $\node{Modal}$ (which ultimately gets its value from $\node{Verb}$
via $\node{Aux}$).  Note however, that the following sentences do {\em
not\/} follow by default from the specifications given at the relevant
nodes:

\[
\begin{array}{l}
\defeq{\npp{Walk}{\val{mor root}}}{\npp{Verb}{\val{mor root}}} \\
\defeq{\npp{Can}{\val{mor past}}}{\npp{Modal}{\val{mor past}}} \\
\defeq{\npp{Aux}{\val{syn type}}}{\npp{Verb}{\val{syn type}}} \\
\end{array}
\]
In each of the above cases, the theory provides an explicit statement
about the value associated with the indicated path at the given node. As
a result the default mechanism is effectively over-ridden.

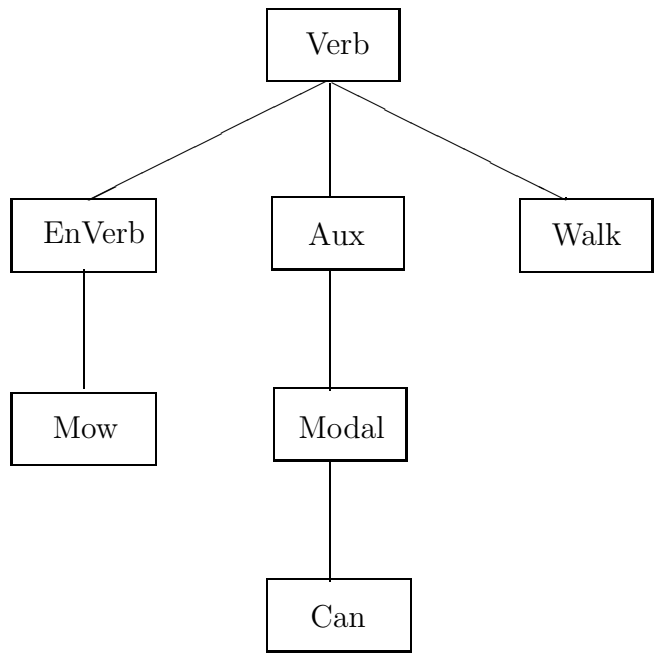
\begin{figure*}
\begin{center}

\setlength{\unitlength}{0.012500in}%
\begingroup\makeatletter\ifx\SetFigFont\undefined
\def\x#1#2#3#4#5#6#7\relax{\def\x{#1#2#3#4#5#6}}%
\expandafter\x\fmtname xxxxxx\relax \def\y{splain}%
\ifx\x\y   
\gdef\SetFigFont#1#2#3{%
  \ifnum #1<17\tiny\else \ifnum #1<20\small\else
  \ifnum #1<24\normalsize\else \ifnum #1<29\large\else
  \ifnum #1<34\Large\else \ifnum #1<41\LARGE\else
     \huge\fi\fi\fi\fi\fi\fi
  \csname #3\endcsname}%
\else
\gdef\SetFigFont#1#2#3{\begingroup
  \count@#1\relax \ifnum 25<\count@\count@25\fi
  \def\x{\endgroup\@setsize\SetFigFont{#2pt}}%
  \expandafter\x
    \csname \romannumeral\the\count@ pt\expandafter\endcsname
    \csname @\romannumeral\the\count@ pt\endcsname
  \csname #3\endcsname}%
\fi
\fi\endgroup
\begin{picture}(268,269)(188,391)
\thinlines
\put(305,470){\makebox(0.1111,0.7778){\SetFigFont{5}{6}{rm}.}}
\put(295,630){\framebox(55,30){}}
\put(297,551){\framebox(55,30){}}
\put(188,550){\framebox(60,30){}}
\put(401,550){\framebox(55,30){}}
\put(298,471){\framebox(55,30){}}
\put(295,391){\framebox(60,30){}}
\put(188,469){\framebox(60,30){}}
\put(321,629){\line( 0,-1){ 48}}
\put(321,551){\line( 0,-1){ 51}}
\put(321,471){\line( 0,-1){ 51}}
\put(218,551){\line( 0,-1){ 50}}
\put(322,629){\line( 2,-1){ 98}}
\put(420,580){\line( 0, 1){  1}}
\put(320,630){\line(-2,-1){100}}
\put(311,641){\makebox(0,0)[lb]{\smash{\SetFigFont{12}{14.4}{rm}Verb}}}
\put(312,561){\makebox(0,0)[lb]{\smash{\SetFigFont{12}{14.4}{rm}Aux}}}
\put(201,562){\makebox(0,0)[lb]{\smash{\SetFigFont{12}{14.4}{rm}EnVerb}}}
\put(414,561){\makebox(0,0)[lb]{\smash{\SetFigFont{12}{14.4}{rm}Walk}}}
\put(205,480){\makebox(0,0)[lb]{\smash{\SetFigFont{12}{14.4}{rm}Mow}}}
\put(308,480){\makebox(0,0)[lb]{\smash{\SetFigFont{12}{14.4}{rm}Modal}}}
\put(313,401){\makebox(0,0)[lb]{\smash{\SetFigFont{12}{14.4}{rm}Can}}}
\end{picture}

\end{center}
\caption{A Lexical Inheritance Hierarchy}\label{hierarchy-fig}
\end{figure*}

In order to understand the use of global (i.e. quoted) inheritance
descriptors it is necessary to introduce \Datr's notion of a {\em global
context\/}.  Suppose then that we wish to determine the value associated with
the path $\path{\val{mor pres}}$ at the node $\node{Walk}$. In this
case, the global context will initially consist of the node/path pair
$\node{Walk}/\path{\val{mor pres}}$.  Now, by default the value
associated with $\path{\val{mor pres}}$ at $\node{Walk}$ is inherited
locally from $\path{\val{mor pres}}$ at $\node{Verb}$.  This, in turn,
inherits {\em globally\/} from the path $\path{\val{mor root}}$. That
is:
\[
\defeq{\npp{Verb}{\val{mor pres}}}{\quoted{\path{\val{mor root}}}}
\]
Consequently, the required value is that associated with $\path{\val{mor
root}}$ at the `global node' $\node{Walk}$ (i.e. the node provided by
the current global context), which is just $\val{walk}$. In a similar
fashion, the value associated with $\path{\val{mor past}}$ at
$\node{Walk}$ is obtained as $\val{walk ed}$ (i.e. the string of atoms
formed by evaluating the specification $\quoted{\path{\val{mor
root}}}\mbox{ }\val{ed}$ in the global context
$\node{Walk}/\path{\val{mor past}}$).

More generally, the global context is used to fill in the missing node
(path) when a global path (node) is encountered. In addition however,
the evaluation of a global descriptor results in the global context
being set to the new node/path pair. Thus in the preceding example,
after the quoted descriptor $\quoted{\path{\val{mor root}}}$ is
encountered, the global context effectively becomes $\node{Walk}$ /
$\path{\val{mor root}}$ (i.e. the path component of the global context
is altered). Note that there is a real distinction between a local
inheritance descriptor of the form $N:P$ and it's global counterpart
$\quoted{N:P}$. The former has no effect on the global context, while
the latter effectively overwrites it.

Finally, the definition of $\node{Verb}$ in the theory of
figure~\ref{datreg-fig} illustrates a use of the `evaluable path'
construct:
\[
\begin{array}{l}
\defeq{\npp{Verb}{\val{mor
form}}}{\quoted{\path{\val{mor}\,\quoted{\path{\val{syn form}}}}}}
\end{array}
\]
This states that the value of $\path{\val{mor form}}$ at $\node{Verb}$
is inherited globally from the path $\path{\val{mor} \cdots}$, where the
dots represent the result of evaluating the global path
$\quoted{\path{\val{syn form}}}$ (i.e. the value associated with
$\path{\val{syn form}}$ in the prevailing global context). Evaluable
paths provide a powerful means of capturing generalizations about the
structure of lexical information.

\section{\Datr\ Models}
\label{sec-models}

To a first level of approximation, the \Datr\ theory of
figure~\ref{datreg-fig} can be understood as a representation of an
inheritance hierarchy (a `semantic network') as shown in
figure~\ref{hierarchy-fig}. In the diagram, nodes are written as
labelled boxes, and arcs correspond to (local) inheritance, or {\em
isa\/} links.  Thus, the node $\node{Can}$ inherits from $\node{Modal}$
which inherits from $\node{Aux}$ which in turn is a $\node{Verb}$. The
hierarchy provides a useful means of visualising the overall structure
of the lexical knowledge encoded by the \Datr\ theory.  However, the
semantic network metaphor is of far less value as a way of thinking
about the \Datr\ language itself. Note that there is nothing inherent in
\Datr\ to ensure that theories correspond to simple {\em isa\/}
hierarchies of the kind shown in the figure. What is more, the \Datr\
language includes constructs that cannot be visualized in terms of
simple networks of nodes connected by (local) inheritance links. Global
inheritance, for example, has a dynamic aspect which is difficult to
represent in terms of static links. Similar problems are presented by
both string values and evaluable paths.  Our conclusion is that the
network metaphor is of primary value to the \Datr\ user. In order to
provide a satisfactory, formal model of how the language `works' it is
necessary to adopt a different perspective.

\Datr\ theories can be viewed semantically as collections of
definitions of partial functions (`nodes' in \Datr\ parlance) that map
paths onto values.  A model of a \Datr\ theory is then an assignment of
functions to node symbols that is consistent with the definitions of
those nodes within the theory. This picture of \Datr\ as a formalism for
defining partial functions is complicated by two features of the
language however. First, the meaning of a given node depends, in
general, on the global context of interpretation, so that nodes do not
correspond directly to mappings from paths to values, but rather to
functions from {\em contexts\/} to such mappings.  Second, it is
necessary to provide an account of \Datr's default mechanism.  It will
be convenient to present our account of the semantics of \Datr\ in two
stages.

\subsection{\Datr\ Interpretations}
\label{sec-interp}

This section considers a restricted version of \Datr\
without the default mechanism.
Section~\ref{sec-default} then shows how implicit
information can be modelled by treating value descriptors as families of
values indexed by paths.

\begin{Def} A \Datr\ {\em interpretation\/} is a triple $I =
(U,\kappa,F)$, where

\begin{enumerate}
\item $U$ is a set;
\item $\kappa$ is a function assigning to each element of the set $(U
\times U^*)$ a partial function from $(U \times U^*)$ to $U^*$.
\item $F$ is a valuation function assigning to each node $N$ and atom
$a$ an
element of $U$, such that distinct atoms are assigned distinct elements.
\end{enumerate}
\end{Def}

Elements of the set $U$ are denoted by $u$ and elements of $U^*$ are
denoted by $v$. Intuitively, $U^*$ is the domain of (semantic)
values/paths.  Elements of the set $C = (U \times U^*)$ are called {\em
contexts\/} and denoted by $c$. The function $\kappa$ can be thought of
as mapping global contexts onto (partial) functions from local contexts
to values.
The function $F$ is extended to paths, so that for $P =
\langle a_1\cdots a_n\rangle$ ($n \ge 0$) we write $F(P)$ to denote $u_1
\cdots u_n \in U^*$, where $u_i = F(a_i)$ for each $i$ ($1 \le i \le
n$).

\begin{figure*}

\[
\begin{array}{lcl}
\denotes{a}_c & = & F(a) \\
\denotes{N:\path{d_1\cdots d_n}}_c & = &
\left\{ \begin{array}{l}
\mbox{ if } v_i = \denotes{d_i}_c \mbox{ is defined for each } i \,
(1 \le i \le n), \mbox{ then } \\
\kappa(c)(F(N), v_1\cdots v_n) \\
\mbox{undefined otherwise}
\end{array} \right. \\
\denotes{\quoted{N:\path{d_1\cdots d_n}}}_c & = &
\left\{ \begin{array}{l}
\mbox{ if } v_i = \denotes{d_i}_c \mbox{ is defined for each } i \,
(1 \le i \le n), \mbox{ then } \\
\kappa(c')(c') \mbox{ where } c' = (F(N),v_1\cdots v_n)  \\
\mbox{undefined otherwise}
\end{array} \right. \\
\denotes{\quoted{\path{d_1 \cdots d_n}}}_c & = &
\left\{ \begin{array}{l}
\mbox{ if } v_i = \denotes{d_i}_c \mbox{ is defined for each } i \,
(1 \le i \le n), \mbox{ then } \\
\kappa(c')(c') \mbox{ where } c = (u,v) \mbox{ and } c' = (u, v_1\cdots
v_n)  \\
\mbox{undefined otherwise}
\end{array} \right. \\
\denotes{\quoted{N}}_c & = & \kappa(c')(c') \mbox{ where } c = (u,v)
\mbox{ and } c' = (F(N), v)
\end{array}
\]
\caption{Denotation function for \Datr\ Descriptors}\label{interp-fig}
\end{figure*}

Intuitively, value descriptors denote elements of $U^*$ (as we
shall see, this will need to be revised later in order to account for
\Datr's default mechanism). We associate with the interpretation $I =
(U, \kappa, F)$  a partial denotation function $\denotes{}:\Desc
\rightarrow (C \rightarrow U^*)$ and write $\denotes{d}_c$ to denote the
meaning (value) of descriptor $d$ in the global context $c$. The
denotation function is defined as shown in figure~\ref{interp-fig}.
Note that an atom always denotes the same element of $U$, regardless of the
context. By contrast, the denotation of an inheritance descriptor is, in
general, sensitive to the global context $c$ in which it appears. Note
also that in the case of a global inheritance descriptor, the global
context is effectively altered to reflect the new local context $c'$.
The denotation function is extended to sequences of value descriptors in
the obvious way. Thus, for $\phi = d_1 \cdots d_n$ ($n \ge 0$), we write
$\denotes{\phi}_c$ to denote $v_1 \cdots v_n \in U^*$ if $v_i =
\denotes{d_i}_c$ ($1 \le i \le n$) is defined (and $\denotes{\phi}_c$ is
undefined otherwise).

Now, let $I = (U, \kappa, F)$ be an
interpretation and $\Theory$ a theory. We will write
$\denotes{\Theory/N}_c$ to denote that partial function from $U^*$ to
$U^*$ given by

\[
\denotes{\Theory/N}_c  =
\bigcup_{\defeq{N:P}{\phi} \in \Theory}\{(F(P),\denotes{\phi}_c)\}
\]
It is easy to verify that $\denotes{\Theory/N}_c$ does indeed denote a
partial function (it follows from the functionality of the theory
$\Theory$).  Let us also write $\denotes{N}_c$ to denote that partial
function from $U^*$ to $U^*$ given by $\denotes{N}_c(v) =
\kappa(c)(F(N),v)$, for all $v \in U^*$.  Then, $I$ {\em models\/}
$\Theory$ just in case the following containment holds for each node $N$
and context $c$:

\[
\denotes{N}_c \supseteq \denotes{\Theory/N}_c
\]
That is, an interpretation is a model of a \Datr\ theory just in case
(for each global context) the function it associates with each node
respects the definition of that node within the theory.

\subsection{Implicit Information and Default Models}
\label{sec-default}

The notion of a model presented in the preceding section is too liberal
in that it takes no account of information {\em implicit\/} in a theory.
For example, consider again the definition of the
node $\node{Walk}$ from the theory of figure~\ref{datreg-fig}, and
repeated below.

\[
\begin{array}{ll}
\node{Walk}: & \path{} == \node{Verb} \\
 & \path{\val{mor root}} == \val{walk}
\end{array}
\]
According to the definition of a model given previously, any model of the
theory of figure~\ref{datreg-fig} will associate with the node
$\node{Walk}$ a function from paths to values
which respects the above definition. This means that for
every global context $c$, the following containment must
hold\footnote{In this and subsequent examples, syntactic objects
(e.g.$\val{walk}$, $\path{\val{mor root}}$) are used to stand for their
semantic counterparts under $F$ (i.e. $F(\val{walk})$, $F(\path{\val{mor
root}})$, respectively).}:


\begin{eqnarray*}
\denotes{\node{Walk}}_c & \supseteq &
  \{ \langle \path{},\denotes{\npp{Verb}{}}_c \rangle, \\
 & & \:\:\: \langle \path{\val{mor root}}, \val{walk} \rangle \}
\end{eqnarray*}

On the other hand, there is no guarantee that a given model will also
respect the following containment:

\begin{eqnarray*}
\denotes{\node{Walk}}_c & \supseteq &
  \{ \langle \path{\val{mor}},\denotes{\npp{Verb}{\val{mor}}}_c \rangle, \\
 & & \:\:\: \langle \path{\val{mor root root}}, \val{walk} \rangle \}
\end{eqnarray*}

In fact, this containment (amongst other things) {\em should\/} hold.
It follows `by default' from the statements made about $\node{Walk}$
that the path $\path{\val{mor}}$ inherits locally from $\node{Verb}$ and
that the value associated with any extension of $\path{\val{mor root}}$
is $\val{walk}$.

There have been a number of formal treatments of defaults in
the setting of attribute-value formalisms
\cite{carp93,bouma92,rbcw92,yr93}. Each of these approaches formalizes a
notion of default inheritance by defining appropriate operations
(e.g. default unification) for combining strict and default information.
Strict information is allowed to over-ride default information where the
combination would otherwise lead to inconsistency (i.e. unification
failure). In the case of \Datr\ however, the formalism does not draw an
explicit distinction between strict and default values for paths. In
fact, all of the information given explicitly in a \Datr\ theory is
strict. The non-monotonic nature of \Datr\ theories
arises from a general, default mechanism which
`fills in the gaps' by supplying values for paths not explicitly
specified in a theory.
More specifically,
\Datr's default mechanism ensures that any path that is not
explicitly specified for a given node will take its definition from the
longest prefix of that path that {\em is\/} specified. Thus, the default
mechanism defines a class of implicit, definitional sentences with paths
on the left that extend paths found on the left of explicit sentences.
Furthermore, this extension of paths is also carried over to paths
occurring on the right. In effect, each (explicit) path is associated
not just with a single value specification, but with a whole family of
specifications indexed by extensions of those paths.

\begin{figure*}
\[
\begin{array}{lcl}
\denotes{\quoted{N:\path{d_1\cdots d_n}}}_c(v) & = &
\left\{ \begin{array}{l}
\mbox{ if } v_i = \denotes{d_i}_c(\epsilon) \mbox{ is defined for each } i
(1 \le i \le n), \mbox{ then } \\
\kappa(c')(c') \mbox{ where } c' = (F(N),v_1\cdots v_n v)  \\
\mbox{undefined otherwise}
\end{array} \right.
\end{array}
\]
\caption{Revised denotation for global node/path pairs}\label{den1-fig}
\end{figure*}

This suggests the following approach to the semantics of defaults in
\Datr. Rather than interpreting node definitions (in a given global
context) as partial functions from paths to values (i.e. of type $U^*
\rightarrow U^*$) we choose instead to interpret them as partial
functions from (explicit) paths, to functions from extensions of those
paths to values (i.e. of type $U^* \rightarrow (U^* \rightarrow
U^*)$). Now suppose that $f:U^* \rightarrow (U^* \rightarrow U^*)$ is
the function associated with the node definition $\Theory/N$ in a given
\Datr\ interpretation.  We can define a partial function $\Delta(f): U^*
\rightarrow U^*$ (the {\em default interpretation\/} of $\Theory/N$) as
follows.  For each $v \in U^*$ set

\begin{eqnarray*}
\Delta(f)(v) & = & f(v_1)(v_2)
\end{eqnarray*}
where $v = v_1v_2$ and $v_1$ is the longest prefix of $v$ such that
$f(v_1)$ is defined. In effect, the function $\Delta(f)$ makes explicit
that information about paths and values that is only
implicit in $f$, but just in
so far as it does not conflict with explicit information provided by
$f$.

In order to re-interpret node definitions in the manner suggested above,
it is necessary to modify the interpretation of value descriptors.
In a given global context $c$, a value descriptor
$d$ now corresponds to a total function $\denotes{d}_c:U^* \rightarrow
U^*$ (intuitively, a function from path extensions to values). For
example, atoms now denote constant functions:

\begin{eqnarray*}
\denotes{a}_c(v) & = & F(a) \mbox{ for all } v \in U^*
\end{eqnarray*}

More generally, value descriptors will denote different values for
different paths. Figure~\ref{den1-fig} shows the revised clause for
global node/path pairs, the other definitions being very similar.  Note
the way in which the `path' argument $v$ is used to extend $v_1 \cdots
v_n$ in order to define the new local (and in this case also, global)
context $c'$. On the other hand, the meaning of each of the $d_i$ is
obtained with respect to the `empty path' $\epsilon$ (i.e. path
extension does not apply to subterms of inheritance descriptors).

As before, the interpretation function is extended to sequences of path
descriptors, so that for $\phi = d_1 \cdots d_n$ ($n \ge o$) we have
$\denotes{\phi}_c(v) = v_1 \cdots v_n \in U^*$, if $v_i =
\denotes{d_i}(v)$ is defined, for each $i$ ($1 \le i \le n$) (and
$\denotes{\phi}_c(v)$ is undefined otherwise).  The definition of the
interpretation of node definitions can be taken over unchanged from the
previous section.  However, for a theory $\Theory$ and node $N$, the
function $\denotes{\Theory/N}_c$ is now of type $U^* \rightarrow (U^*
\rightarrow U^*)$.  An interpretation $I = (U,\kappa,F)$ is a {\em
default model\/} for theory $\Theory$ just in case for every context $c$
and node $N$ we have:
\begin{eqnarray*}
\denotes{N}_c & \supseteq & \Delta(\denotes{\Theory/N}_c)
\end{eqnarray*}

As an example, consider the default interpretation of the definition of
the node $\node{Walk}$ given above. By definition, any default model of
the theory of figure~\ref{datreg-fig} must respect the following
containment:

\begin{eqnarray*}
\denotes{\node{Walk}}_c & \supseteq &
 \Delta(\{\langle \path{}, \lambda v. \denotes{\npp{Verb}{}}_c(v)
\rangle, \\
 & & \:\:\:\:\:\:\:\langle \path{\val{mor root}}, \lambda
v. \val{walk} \rangle \}
\end{eqnarray*}
{}From the definition of $\Delta$, it follows that for any path $v$, if
$v$ extends $\path{\val{mor root}}$, then it is mapped onto the value
$\val{walk}$, and otherwise it is mapped to the value given by
$\denotes{\npp{Verb}{}}_c(v)$. We have the following picture:
\begin{eqnarray*}
\denotes{\node{Walk}}_c & \supseteq &
 \{
 \langle \path{},
 \denotes{\npp{Verb}{}}_c(\path{}) \rangle, \\
 & & \:\:\:
 \langle \path{\val{mor}},
 \denotes{\npp{Verb}{}}_c(\path{\val{mor}}) \rangle, \\
 & & \:\:\:\ldots \\
 & & \:\:\:
 \langle \path{\val{mor root}}, \val{walk} \rangle, \\
 & & \:\:\:
 \langle \path{\val{mor root root}}, \val{walk} \rangle, \\
 & & \:\:\: \ldots \}
\end{eqnarray*}

The default models of a theory $\Theory$ constitute a proper subset of
the models of $\Theory$: just those that respect the default
interpretations of each of the nodes defined within the theory.

\section{Conclusions}
\label{sec-conc}

The work described in this paper fulfils one of the objectives of the
\Datr\ programme: to provide the language with an explicit, declarative
semantics. We have presented a formal model of \Datr\ as a language for
defining partial functions and this model has been contrasted with an
informal view of \Datr\ as a language for representing inheritance
hierarchies.  The approach provides a transparent treatment of \Datr's
notion of (local and global) context and accounts for \Datr's default
mechanism by regarding value descriptors (semantically) as families of
values indexed by paths.

The provision of a formal
semantics for \Datr\ is important for several reasons. First, it
provides the \Datr\ user with a concise, implementation-independent
account of the meaning of \Datr\ theories.  Second, it serves as a
standard against which other, operational definitions
of the formalism can be judged.  Indeed, in the absence of such a
standard, it is impossible to demonstrate formally the correctness of
novel implementation strategies (for an example of such a strategy, see
\cite{la94}).  Third, the process of formalisation itself aids our
understanding of the language and its relationship to other
non-monotonic, attribute-value formalisms.  Finally, the semantics
presented in this paper provides a sound basis for subsequent
investigations into the mathematical and computational properties of
\Datr.

\section{Acknowledgements}

The author would like to thank Roger Evans, Gerald Gazdar, Bill Rounds
and David Weir for helpful discussions on the work described in this
paper.


\end{document}